\documentclass[sigconf]{acmart}

\usepackage{tikz, pgf, subcaption}
\usepackage{hyperref}
\usepackage{float}
\usepackage{pgfplots}
\usepackage{minted}
\usepackage{caption}
\usepackage{mdframed}
\usepackage{multirow}
\usetikzlibrary{calc}
\pgfplotsset{compat=1.18}

\AtBeginDocument{%
  }

\copyrightyear{2026}
\acmYear{2026}
\setcopyright{cc}
\setcctype{by}
\acmConference[GECCO '26]{Genetic and Evolutionary Computation Conference}{July 13--17, 2026}{San Jose, Costa Rica}
\acmBooktitle{Genetic and Evolutionary Computation Conference (GECCO '26), July 13--17, 2026, San Jose, Costa Rica}
\acmDOI{10.1145/3795095.3805109}
\acmISBN{979-8-4007-2487-9/2026/07}

\begin{document}

\title{Optimizing Chlorination in Water Distribution Systems via Surrogate-assisted Neuroevolution}

\author{Rivaaj Monsia}
\orcid{0009-0001-8783-2664}
\affiliation{%
  \institution{UT Austin}
  \city{Austin}
  \state{Texas}
  \country{USA}
}
\email{rivaaj@utexas.edu}

\author{Daniel Young}
\orcid{0009-0009-9319-7554}
\affiliation{%
  \institution{UT Austin \& Cognizant AI Lab}
  \city{Austin, TX}
  \state{San Francisco, CA}
  \country{USA}
}
\email{danyoung@utexas.edu}

\author{Olivier Francon}
\orcid{0009-0009-0006-4913}
\affiliation{%
  \institution{Cognizant AI Lab}
  \city{San Francisco}
  \state{California}
  \country{USA}
}
\email{olivier.francon@cognizant.com}

\author{Risto Miikkulainen}
\orcid{0000-0002-0062-0037}
\affiliation{%
 \institution{UT Austin \& Cognizant AI Lab}
 \city{Austin TX}
 \state{San Francisco, CA}
 \country{USA}
}
\email{risto@cs.utexas.edu}

\renewcommand{\shortauthors}{Monsia, et. al.}

\begin{abstract}
Ensuring the microbiological safety of large, heterogeneous water distribution systems (WDS) typically requires managing appropriate levels of disinfectant residuals including chlorine. WDS include complex fluid interactions that are nonlinear and noisy, making such maintenance a challenging problem for traditional control algorithms. This paper proposes an evolutionary framework to this problem based on neuroevolution, multi-objective optimization, and surrogate modeling. Neural networks were evolved with NEAT to inject chlorine at strategic locations in the distribution network at select times. NSGA-II was employed to optimize four objectives: minimizing the total amount of chlorine injected, keeping chlorine concentrations homogeneous across the network, ensuring that maximum concentrations did not exceed safe bounds, and distributing the injections regularly over time. Each network was evaluated against a surrogate model, i.e.\ a neural network trained to emulate EPANET, an industry-level hydraulic WDS simulator that is accurate but infeasible in terms of computational cost to support machine learning. The evolved controllers produced a diverse range of Pareto-optimal policies that could be implemented in practice, outperforming PPO, a standard reinforcement learning method. The results thus suggest a pathway toward improving urban water systems, and highlight the potential of using evolution with surrogate modeling to optimize complex real-world systems.
\vspace*{-2ex}
\end{abstract}




\maketitle

\section{Introduction}

\par
Of the various challenges in supporting urban growth, the implementation and maintenance of water distribution systems (WDS) represents a particularly critical issue. From 2010 to 2030, urban land use has been estimated to grow by about 1.5 million km$^2$ \cite{seto}. Thus, robust short and long-term planning of WDS is essential for development; indeed, access to clean water is recognized as one of the United Nations' 17 Sustainable Development Goals \cite{United}.
\par
Importantly, these distribution systems ensure chemical and microbiological safety of drinking water via various methods for chemical disinfection. Typically, chlorine-based disinfectants are injected into water utilities to control for microbial activity; this approach, specifically the use of chlorine, is relatively easy to implement and has a low cost \cite{Grayman_2008}. 
\par
In order to maintain compliance in various jurisdictions (including the US), chlorine residuals need to be limited \cite{Roth_Cornwell_2018}. However, it is challenging to maintain appropriate levels due to various reasons, including water age and consumer demands and interactions \cite{Fisher_Kastl_Sathasivan_2011, Hua_Vasyukova_Uhl_2015, Li_Feng_Li_Yang_Zhang_2022}. Moreover, byproducts build up from reactions with bacteria and microbial derivatives \cite{Zhou_Bian_Yang_Fu_2023, Shah_Mazhar_Ahmed_Lew_Khalil_2024}. Elevated concentration of these byproducts, typically categorized by their aliphatic or aromatic molecular nature, have been shown to be correlated with varying adverse ailments, including bladder, liver, and colon cancers, cardiovascular diseases, and dysregulated prenatal development \cite{Kalita_Kamilaris_Havinga_Reva_2024}. Ensuring spatio-temporally even disinfectant residuals across WDS remains a challenging task for large-scale implementation and deployment.
\par
Unfortunately, experimentation and benchmarking cannot be done in active WDS systems because of their critical role in existing infrastructure. Instead, WDS must be computationally simulated in order to provide an environment that allows proper engineering and optimization. Moreover, human control is practically infeasible due to the complex topology of WDNs alongside the dynamic, nonlinear interactions of fluid, control, and species reaction dynamics.
\par
At the same time, AI agents and reinforcement learning (RL) have shown much promise in control systems \cite{Sarker_2021, Schöning_Riechmann_Pfisterer_2022}. In general, the need for reinforcement learning algorithms has exploded as opportunities for new agents and agentic systems with real-world applications have been realized. Evolutionary optimization is particularly well-suited for such tasks. Leveraging evolutionary principles in multi-objective, noisy, and deceptive fitness landscapes has proven useful in producing diverse, creative solutions compared to standard RL algorithms.

\vspace*{0.5ex}
\hspace*{-3ex}
\begin{minipage}{\columnwidth}
\hspace*{3ex}In this paper, an evolutionary framework is proposed for developing neural networks to control chlorine injections based on concentration and flow measurements within a WDS. The approach takes advantage of the ESP framework ~\cite{Francon_Gonzalez_Hodjat_Meyerson_Miikkulainen_Qiu_Shahrzad_2020} to successively (1) train a meta-surrogate model of the WDS simulator, (2) evolve NEAT (Neuroevolution of Augmenting Topologies) ~\cite{stanley2002evolving} networks against the surrogate, and (3) expand the surrogate's training data by evaluating the best evolved agents on the hydraulic simulator. Evolution is carried out through multi-objective optimization in which the agents are evaluated on four objectives, which are vital to robust, parsimonious chlorination control systems.
\end{minipage}
\par
In extensive experiments with the EPANET simulator ~\cite{EPA, epyt-flow}, the proposed framework produced favorable objective tradeoffs compared to a breadth of policies. Perpetual fine-tuning of the surrogate model resulted in a virtuous cycle where the agent implicitly regularizes the surrogate, and the surrogate enables increased exploration of the reward landscape.
\par
In the future, the surrogate can be further improved by incorporating more physical constraints. Spatial heuristics can be encoded into the agents, and agent communication can be established to expedite learning.
\par
Thus, the proposed framework provides a new avenue in which chlorination control systems may be built. In turn, these systems could be leveraged to support urban and infrastructural growth by ensuring the microbiological safety of drinking water across the world.
\par
All code is available at \texttt{\href{https://github.com/rivmons/moea-esp-ccontrol}{rivmons/moea-esp-ccontrol}}.

\vspace*{-1ex}
\section{Background}
The structure of WDS is first described, followed by a review of prior work on modeling and optimizing them, leading to the approach in this paper. 

\begin{figure}
    \centering
    \includegraphics[width=\linewidth]{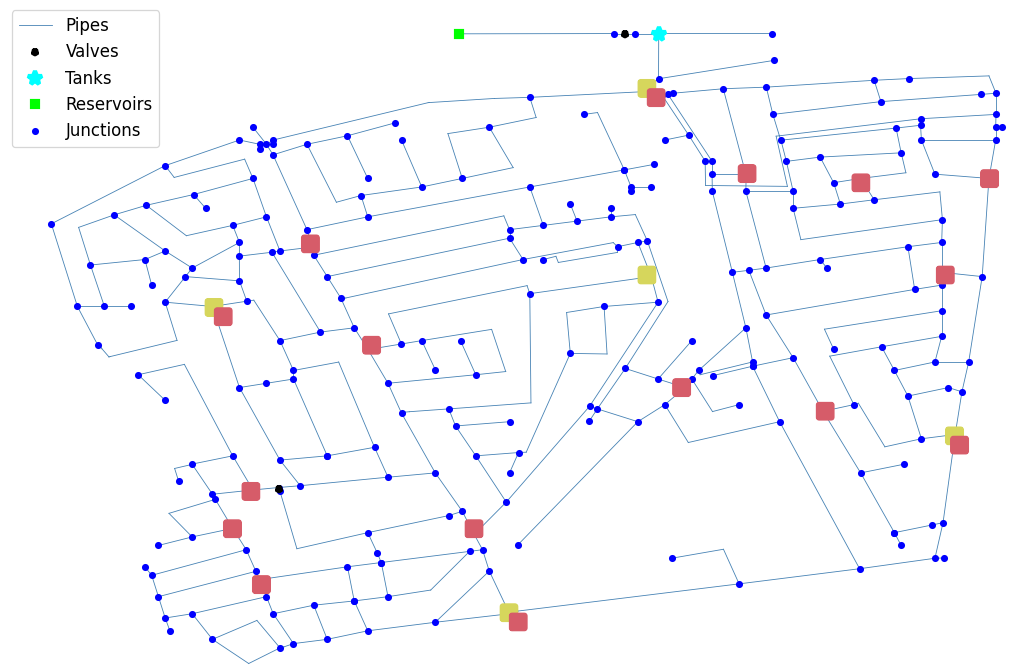}
    \caption{Topology of the water distribution network utilized in this paper. Red and yellow squares denote sensor and injection nodes, respectively; reservoirs indicate infinite water sources. The topology defines the optimization task and the challenges in it. In this case, the large diameter introduces long-term effects and high-density nonlinear effects, which the surrogate model needs to learn and the controller needs to optimize.
    }
    \label{fig2}
    \Description{A representation of a water distribution network as a graph, with junctions as nodes and pipes as edges. Additionally, valves to control flow are interspersed across the WDS, and a reservoir and tank at the top of the network represent the water source.}
    \vspace*{-2ex}
\end{figure}

\vspace*{-1ex}
\subsection{Water Distribution Systems}
A water distribution system is a large network of pipes that distributes drinking water from sources to consumer nodes (Figure~\ref{fig2}). Inherently, it  can be represented as a graph 
$\mathcal{G} = (V, E)$,
where $V$ is the set of nodes and $E$ is the set of edges corresponding to simple pipes, pumps, and valves.  
Each pipe, $e \in E$ has geometric and material parameters such as length $l_e$, diameter $\delta_e$ and roughness coefficient $c_e$, where $l_e, \delta_e, c_e \in \mathbb{R}_+$ (the set of reals $\ge 0$). The entire WDS can be temporally modeled by water demands $\mathbf{d(t)} \in \mathbb{R}_+^{|V|}$, flows $\mathbf{q(t)} \in \mathbb{R}^{|E|}$, and heads $\mathbf{h(t)} \in \mathbb{R}_+^{|V|}$. The head at a certain node $v \in V$, is a measurement of liquid pressure with respect to fluid height: $d_v = p_v + \epsilon_v$, where $p_v$ is the fluid pressure at node $v$ and $\epsilon_v$ is the elevation \cite{Kyriakou_Demetriades_Vrachimis_Eliades_Polycarpou_2023}. In turn, the head is used to calculate headloss, or the loss of energy and fluid pressure due to friction and other factors.
\par
Water distribution systems consist of two tightly coupled components:
\emph{hydraulics}, which governs the movement of fluid through the network, and
\emph{water quality}, which governs how disinfectants and contaminants (species) are transported and react with other chemicals and microorganisms \cite{Bhave_1991}. These two components are interdependent, and the interactions between them dictate the need for multi-objective considerations for WDS control systems.

\paragraph{Hydraulic Dynamics}
The hydraulic state is simply defined by the pressure head at each node $\mathbf{h_v(t)}$ and the flow $\mathbf{q_e}$ in each pipe. Fundamentally, if two connected nodes have different pressure heads, water flows down the pressure gradient. When demand at consumer nodes changes, the hydraulic state must change due to increased reservoir supply and, therefore, altered flows and heads. 

\par
The hydraulic state of a system is defined by three conservation principles \cite{aichall}: the conservation of flow, mass, and energy:
\allowdisplaybreaks
\begin{align}
    q_{vu}(t) &= -q_{uv}(t) \label{flow} \\
    \sum_{v \ \in \ \mathcal{N}(u)} q_{vu}(t) &= d_u(t) \label{mass} \\
    h_v(t) - h_u(t) &= \Delta h_{vu}(q_{vu}(t)). \label{hazen}
\end{align}


\par
Hydraulics are integral in maintaining spatiotemporally even chlorine concentrations. Once injected, chlorine is transported across pressure gradients and undergoes natural decay. Because water velocities depend on system-wide demands, chlorine spikes propagate at varying speeds. Thus, flow measurements are necessary to inform a controller on the movement of fluid through a network \cite{aichall, Bhave_1991}.

\paragraph{Water Quality}
In addition to hydraulics, the WDS constitutes a quality state in which the concentration of species like chemicals and bacteria is tracked at each timestep. Typical species include chlorine, fluoride, or various bacteria that cause human disease. Formally, the quality state at time $t$ is defined as 
$
\mathbf{c(t)} \in \mathbb{R}_{\geq 0}^{|V| \times |S|},
$
where each node has a concentration value for each species $s \in S$.

\par
Just like hydraulic states, the quality state is governed by three principles, namely advection, diffusion, and reaction. Advection is the translational transport of species due to flow. Diffusion is the radial movement of species due to concentration gradients. Lastly, reaction is, just as the name implies, the transformation or degradation of various species to other species. Typically, dynamics are dominated by advection and reaction \cite{aichall}. Chlorine concentration is generally modeled as a decay process with respect to time. As such, water age and flows are essential in chlorination control systems.

\vspace*{-1ex}
\subsection{Related Work}
\par
Significant prior work exists on the optimization of water distribution network tasks, from service-pressure control to optimization of hydroelectric energy production \cite{Creaco_Campisano_Fontana_Marini_Page_Walski_2019}. The utility of digital twins in modeling these networks has also been substantiated \cite{Ashraf_Strotherm_Hermes_Hammer_2024}. Of note are a variety of online control algorithms that make it possible to maintain chlorine concentration by modeling chemical transport through multi-species reaction dynamics \cite{Elsherif_Taha_Abokifa_2024}. Specifically, Elsherif \textit{et. al.} \cite{Elsherif_Taha_Abokifa_2024} built a model-predictive control algorithm to determine the magnitude of chlorine injections, aligning with the surrogate-based modeling of the hydraulic system in this paper.

The breadth of these problems is prominent enough to prompt competitions and challenges that focus on WDS optimization. In particular, the AI for Drinking Water Chlorination Challenge was held at the International Joint Conference on Artificial Intelligence (IJCAI) in 2025, in which multiple successful approaches to chlorination control were presented from rule-based systems to deep RL solutions \cite{aichall}.
\par 
Numerous machine learning (ML) solutions have been applied to disinfectant control and WDS-adjacent optimization tasks. For example, Wu \textit{et. al.} used graph convolutional networks (GCN) to model the spatial correlations inherent in modeling water demand, while Fan \textit{et al.} approached the problem of leakage detection via autoencoders \cite{Wu_Wang_Liu_Yu_Wu_2023, Fan_Zhang_Yu_2021}. 
There has also been significant work in simulating the often computationally inefficient hydraulic simulators through neural-network-based estimators. Various architectures and training schematics have been proposed, including a graph neural network (GNN), which compounds physics-informed constraints to predict WDS species concentrations alongside flows~\cite{Ashraf_Strotherm_Hermes_Hammer_2024}.

The approach in this paper uses a similar neural-network-based estimator, but, importantly, does not take into account spatial correlations of the WDS or enforce physical constraints explicitly. While this approach requires more data to learn the same constraints, it allows control policies to guide the training of the surrogate model. Thus, diversity and exploration are central to this approach, taking advantage of the creativity that evolutionary methods provide.

\vspace*{-1ex}
\section{Method}
\label{sc:method}

\par
The method of simulating WDS is first reviewed, followed by a description of the data produced from it. Then, each component of the evolutionary optimization approach is described, including Evolutionary Surrogate-Assisted Prescription~\cite[ESP;][]{Francon_Gonzalez_Hodjat_Meyerson_Miikkulainen_Qiu_Shahrzad_2020}, surrogate distillation, Neuroevolution of Augmenting Topologies~\cite[NEAT;][]{stanley2002evolving}, and multi-objective optimization.            


\vspace*{-1ex}
\subsection{WDS Emulation via EPANET}
The hydraulic simulator utilized to model water distribution networks in this paper was generated with \texttt{EPyT} \cite{epyt-flow}, a Python interface to the EPANET hydraulic and water quality simulator. EPANET utilizes a Lagrangian time-based approach to track the movement of discretized units of fluid \cite{EPA, Kyriakou_Demetriades_Vrachimis_Eliades_Polycarpou_2023}. \texttt{EPyT} provides simulations of water distribution systems on top of EPANET by repeatedly solving a system of differential/algebraic equations. At each timestep, the simulator updates flows, pressures, and chemical concentrations based on consumer demand, system parameters affecting transportation and reaction dynamics, and any control actions applied to the system (e.g.\ chlorine booster injections). From the simulation, trajectories of chlorine concentrations, flow rates, and other state variables, and the corresponding control inputs and sensor readings specific to each WDS can be obtained for surrogate training. Ultimately, a surrogate was trained to estimate the simulator state.




\vspace*{-1.5ex}
\subsection{Data Generation}
\par
The data used to initialize the hydraulic simulator came from the AI for Drinking Water Chlorination Challenge at IJCAI 2025 \cite{aichall}. The simulations were based on a WDS with a specific fixed topology shown in \autoref{fig2}.
Entering water has a random chlorine concentration from 0.4 to 0.6 mg/L) and contains a variable concentration of organic matter that can react with chlorine.
\par
To implement a control policy, the network contains five nodes that act as chlorine booster stations. An agent can decide to inject chlorine in these stations in the range $[0, 10000]$ mg/L. To inform the policy, chlorine levels at 17 monitoring nodes alongside flow measurements at two pipes are observed. The simulations run for three days at a resolution of five minutes/timestep. While each scenario utilizes the same topology, different system-wide parameters governing reaction dynamics and fluid transport define different scenarios. 

At each timestep in the simulation, water demands vary, flows and heads are recalculated, reaction dynamics occur, and a controller can inject chlorine at the booster stations. Water demands and organic concentrations vary over time (typically lower at nighttime), season (higher in the summer), and household.


\par
In addition to the normal operation, contamination events occur occasionally in the network. In such an event, an abnormally high amount of pathogens or organic matter is injected into the network. The time and place in which these events occur cannot be sensed by the controller directly, but it has to learn to mitigate them based on its usual observations.

Controllers must optimize five objectives: infection risk, chlorine bound violations, total chlorine used (cost), maintenance of spatially-even chlorine concentrations (fairness), and similar injection magnitudes at adjacent timesteps (smoothness). In this paper, infection risk is not directly optimized. This objective is dependent
on contamination events that occur randomly. Thus, it is not possible to calculate this risk deterministically for each timestep, and therefore it is not included in the optimization scheme. However, it is possible to assess how well the model implicitly minimizes infection risk by observing how well it handles contamination events. Infection risk is therefore included as the fifth objective when evaluating the solutions in the end.
The health and cost objectives are in conflict, and therefore solutions represent tradeoffs between them. See \autoref{R-app} for a detailed description of these objectives.

\subsection{Evolutionary Surrogate-Assisted Prescription (ESP)}
\par

The ESP framework is comprised of two models, a prescriptor $P_s$ that prescribes actions $A$ from observations $O$ \eqref{eq:1} and a predictor $P_d$ that models a reward landscape and predicts a reward $R$ from an observation-action $C, A$ pair \eqref{eq:2}:
\begin{align}
    P_s(O) = A \label{eq:1}\\
    P_d(C, A) = R. \label{eq:2}
\end{align}

Instead of utilizing an explicit reward function, which may be costly, a surrogate model (or predictor) may be deployed, speeding up learning. In many benchmark tasks, ESP converges faster than standard RL methods (PPO, DQN) with a much lower variance and cost (or regret) \cite{Francon_Gonzalez_Hodjat_Meyerson_Miikkulainen_Qiu_Shahrzad_2020}. 

\begin{figure}
    \centering
    \includegraphics[width=0.7\linewidth]{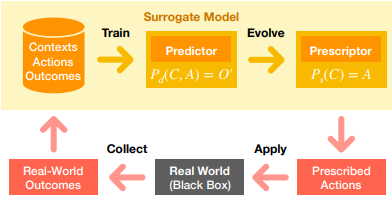}
    \caption{The outer loop of the Evolutionary Surrogate-assisted Prescription (ESP) algorithm \cite{Francon_Gonzalez_Hodjat_Meyerson_Miikkulainen_Qiu_Shahrzad_2020}. The predictor and prescriptor are co-optimized to guide learning of both models synergistically, encouraging innovation and regularizing the resulting policies.}
    \label{fig4}
    \Description{The workflow of the ESP approach. The predictor is trained on data, then used to evolve prescriptors. The prescriptors are applied, and data is collected to re-train the surrogate.}
\vspace*{-1ex}
\end{figure}

ESP takes form as an iterative algorithm until agents/prescriptors reach convergence (see \autoref{fig4}): 

\begin{enumerate}
    \item Collect initial data from the hydraulic simulator by injecting random amounts of chlorine.
    \item Train a surrogate on said data via gradient descent.
    \item Evolve NEAT-based prescriptors (or agents) via NSGA-II \cite{nsga-2}, using the surrogate instead of the explicit reward function.
    \item Apply the best prescriptor(s) on the hydraulic simulator and collect new data.
    \item Repeat from Step~2 until the prescriptor population converges.
\end{enumerate}

In the WDS use case, the initial data is generated once and used to train the initial surrogate. Afterwards, the surrogate is further trained on a mixture on new data from the best evaluated agents and the initial data. Unlike typical ESP, instead of using the predictor to model the reward landscape directly, in the WDS application, it models the hydraulic and quality states of the WDS. The rewards are then calculated deterministically by caching the agents' actions and the surrogate's estimation of the state (see \autoref{R-app} for details).
 
The reason for this design is that the simulation of the hydraulics and quality is sequential in nature and therefore computationally expensive. By instead learning a meta-simulator, the cost of evaluating agents reduces significantly. Moreover, the prediction of the state variables can be gradually improved as the prescription agents improve. The details of this process are described next.

\subsection{Constructing Surrogate Models through Knowledge Distillation}

The surrogate, i.e.\ a uni-directional LSTM denoted as $g_\theta$, must predict the next-step hydraulic and water quality state $x_t = [f_t, o_t, a_t]$, i.e\ flows, observations, and actions at timestep $t$, over a causal context window $C_t$. To reduce long-horizon error, the model employs state increments $\Delta_t = y_{t+1} - y_t$ rather than raw values. The next-step prediction is then $\hat{y}_{t+1} = y_t + \hat{\Delta}_t$. Naturally, the surrogate operates on multi-step forecasts by autoregressively feeding predictions into $C_t$. 
\par
In preliminary experiments, learning long-term dependencies turned out to be difficult with $g_\theta$. Thus, a teacher-student framework is employed. The teacher $T_\phi$ is a bi-directional LSTM with additional lookahead context $\tilde{C}_t$, producing non-causal targets $\hat{\Delta}_t^{(T)}$, while the student $S_\theta$ is causal and trained via a hybrid loss combining four variable loss terms, including a ground-truth and a distillation term.
\par
Training data is of the form $\{(f_t, o_t, a_t, o_{t+1})\}$ and collected from multiple EPANET scenarios. Both models are trained with early stopping and the student $S_\theta$ is fine-tuned on trajectories from evolved policies. Additional information about the teacher-student scheme and surrogate training can be found in \autoref{S-app}.

\subsection{Constructing Prescriptors through NEAT}

After training the surrogate models, the second component of ESP is to construct the prescriptors that will control chlorine injections in a WDS. 
\par
To generate these agents, NEAT \cite[Neuroevolution of Augmenting Topologies;][]{stanley2002evolving} is applied, a population-based method that simultaneously evolves both parameters and topology of an artificial neural network. NEAT requires no gradient information, complexifies parsimoniously, and establishes diversity through speciation. In prior work, NEAT has been applied successfully to several control tasks, and it is therefore a good approach for WDS control as well.
\par
In the WDS control tasks, training contexts $C_t$ are sampled, and the agent needs to generate control actions for time $t$: 
\[r_{\psi}(o_t, f_t) = \hat{a}_t.\]
During evolution, the surrogate takes as input $x'_t= [o_t, f_t, \hat{a}_t]$ to generate a new state $\hat{y}_{t+1} = [\hat{f}_{t+1},\hat{o}_{t+1}]$. This new state left-shifts the context vector $C_t$ to $C_{t+1}$, and this process is repeated in an auto-regressive fashion to generate state predictions that may be used to calculate the fitness for each agent. Standard evolution operations of crossover and mutation are then used to create new agents.

\subsection{Multi-objective Optimization and Curriculum Learning}
In order to optimize the four WDS quality metrics simultaneously, a multi-objective optimization framework is needed. To ensure that each objective is optimized, curriculum evolution is used: starting with two objectives, the remaining objectives are gradually introduced.
\par
\subsubsection{Multi-objective optimization}
NEAT agents are evolved to optimize four objectives: bound violations, fairness, cost, and smoothness using the NSGA-II framework \cite{nsga-2}. NSGA-II sorts individuals into successive layers of Pareto fronts and focuses reproduction on the primary layers, i.e.\ the most useful tradeoffs between objectives. Reproduction occurs via binary tournament mating selection.
\par
For the final evaluation, candidates that represent the best tradeoffs are selected from the Pareto front and evaluated accurately in the EPANET hydraulic simulator.

\subsubsection{Curriculum Learning}
\label{sc:curr_3}
\par
To ensure that the agents learn meaningful tradeoffs, curriculum learning is incorporated into the evolution scheme. Models and agents are trained on examples and constraints of increasing difficulty \cite{soviany2022curriculumlearningsurvey}. This approach mirrors human learning that starts from simple, basic skills (e.g.\ addition) and progresses to more complex ones step-by-step (e.g.\ calculus).
\par
This approach was motivated by preliminary experiments. When the agents were evolved with all objectives simultaneously, unintuitive tradeoffs often emerged. In contrast, when objectives were included incrementally, the population of neural networks finds niches that become gradually more refined. In preliminary experiments, the order of increasing difficulty was found to be bound violations, fairness, smoothness, and cost. Starting from the first two together, this order was used as the default in the experiments.

Thus, curriculum learning leverages the existence of multiple objectives to reliably guide the search towards the best tradeoffs.

\section{Results}

In this section, the results of applying the above methods to chlorination control are presented. The outcomes of surrogate training are first reviewed, followed by qualitative and quantitative performance of prescriptor evolution, and the effects of surrogate fine-tuning. The details of the experimental setup are in Appendices~\ref{S-app} and~\ref{N-app}.

\subsection{Surrogate Training}

Two metrics and five versions of the surrogate loss function were evaluated and used to predict how the simulation unfolds over time.


\subsubsection{Evaluation Metrics}
The performance of the student model was evaluated with/without knowledge distillation and with various loss terms. Mean Squared Error (MSE) and Mean Absolute Error (MAE), common in numerical optimization tasks, were used to quantify the error between student and teacher predictions against true chlorine concentrations.

Moreover, the auto-regressive performance of the surrogate was assessed by quantifying its error over $n$ timesteps. This \emph{Horizon MSE} \eqref{mse_h} was calculated over four horizons consisting of 5, 10, 20, and 50 steps. This range corresponds to real-time horizons from 25 minutes to 250 minutes. Formally, the Horizon MSE is defined as
\begin{equation}
\label{mse_h}
\mathrm{MSE}_{\text{horizon}}
= 
\frac{1}{N}
\sum_{i=1}^{N}
\sum_{h=1}^{H}
\left\lVert 
\mathbf{y}_{i}^{(t+h)} - \hat{\mathbf{y}}_{i}^{(t+h)}
\right\rVert_2^{2},
\end{equation}
where $N$ is the number of auto-regressive sequences, $y_{i}^{(h)}$ is the ground-truth observation at step $t+h$ and $\hat{y}_{i}^{(t+h)}$ is the model prediction at $t + h$.

\subsubsection{Loss Function Design} \label{lfabl}
Five different versions of the loss function were evaluated with different coefficients $\lambda_s$, $\lambda_h$, $\lambda_f$, and $\lambda_r$ for the hard, soft, feature, and rollout loss (see \autoref{S-app} for details). Coefficients for each term were set heuristically, with hard and soft losses weighed most heavily. Teacher and student models were trained over 50 epochs. The validation split was 10\% of contexts from each scenario, and the batch size was 256.

\begin{table*}[h]
\centering
\caption{Surrogate Prediction Accuracy across Various Loss-function Versions.
Metrics are averaged over the validation set. Horizon MSE is broken into 
four prediction ranges. The versions are named according to their nonzero loss coefficients. The best predictions were achieved with HSR and All, suggesting that the rollout loss is most important to include in the loss. HSR performed the best and was therefore used in the experiments.
}
\begin{tabular}{lccccccccccc}
\toprule
{\textbf{Version}} 
& \multicolumn{4}{c}{\textbf{Loss Coefficients}} 
& \multicolumn{4}{c}{\textbf{Horizon MSE}} 
& \multicolumn{3}{c}{\textbf{Single-Step}} \\
\cmidrule(lr){2-5} \cmidrule(lr){6-9} \cmidrule(lr){10-12}
& $\lambda_s$ & $\lambda_h$ & $\lambda_f$ & $\lambda_r$
& 5 & 10 & 20 & 50
& MSE & Soft-MSE & MAE \\
\midrule

H (Uni) & 0.0 & 1.0 & 0.0 & 0.0 
& 42.2 & 59.8 & 105.1 & 253.7
& \textbf{0.935} & 0.884 & \textbf{0.628} \\


HS & 0.5 & 0.5 & 0.0 & 0.0 
& 42.5 & 67.2 & 120.5 & 271.2
& 0.941 & 0.888 & 0.630 \\

HSF & 0.5 & 0.5 & 0.1 & 0.0 
& 43.0 & 61.7 & 101.0 & 190.2 
& 0.938 & 0.850 & 0.630 \\

HSR & 0.5 & 0.5 & 0.0 & 0.2 
& \textbf{33.3} & \textbf{34.0} & \textbf{37.2} & \textbf{42.9} 
& 0.949 & \textbf{0.782} & 0.634 \\

All & 0.5 & 0.5 & 0.1 & 0.2 
& 33.9 & 34.6 & 37.7 & 43.1
& 0.950 & 0.804 & 0.634 \\

\bottomrule
\end{tabular}\\[2ex]
\label{table1}
\end{table*}

\par
The results in \autoref{table1} indicate that single-step state prediction ($\hat{y}_{t+1}$) is relatively similar across all loss configurations.
Nevertheless, insightful observations can be made. For instance, 
despite more terms, HSR significantly outperformed HS, indicating that the rollout loss term is co-optimized with the soft loss term. This result was to be expected since teacher predictions are based on lookahead context $\tilde{C}_t$ and therefore include a representation of delayed state dynamics. 
\par
The most consequential term is the rollout loss, which significantly decreases Horizon MSE in the HSR and All configurations. It is therefore beneficial to include this term in distillation for the current use case or even in the development of pure model predictive control (MPC) algorithms, which enforce physical constraints \cite{Ashraf_Strotherm_Hermes_Hammer_2024}.


\subsubsection{Single-Step $\Delta_t$ Prediction}
To evaluate surrogate performance, contiguous context sequences from the validation set were sampled to observe, graphically, how well the teacher and student follow $y_t$ via $\hat{\Delta}_t$ and the simple relation defined in \autoref{delta:eq}.
\par
\begin{figure}
    \centering
    \includegraphics[width=\linewidth]{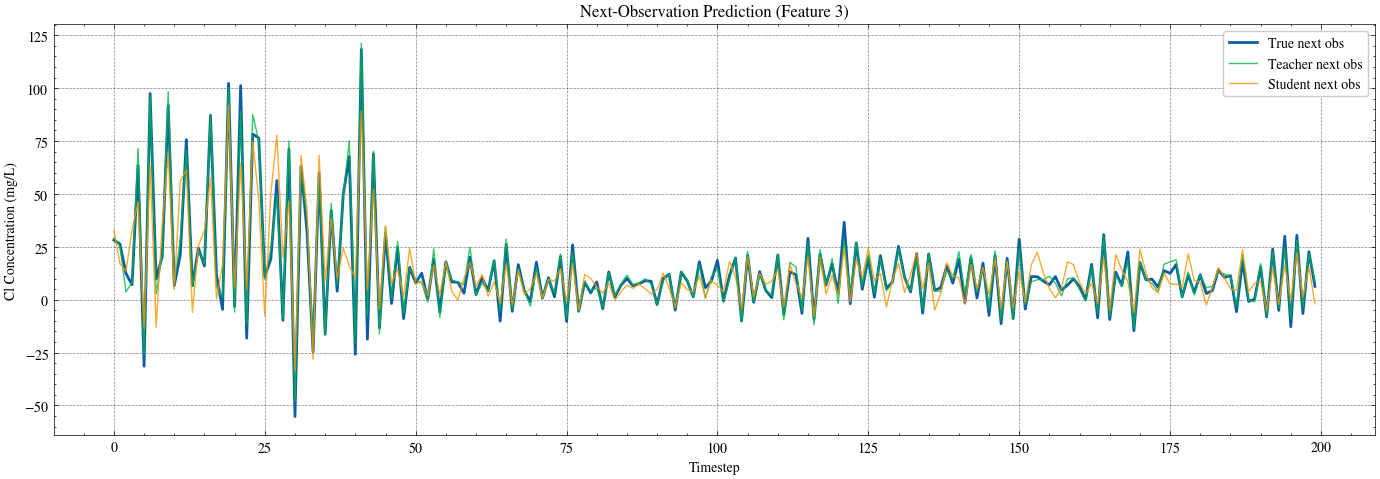}
    \caption{Comparison of $T_{\phi}$ vs. $S_{\theta}$ vs.\ true values for $o_{t+1}$/$\hat{o}_{t+1}$. Low error in $S_{\theta}$ predictions despite large variations in concentration suggests that the distillation from $T_{\phi}$ is reliable, and the resulting surrogate model can be used to optimize the controller.}
    \Description{3 figures illustrating the predictions of the teacher and student model against the true chlorine concentration. The top figure follows scaled delta increment predictions, the middle shows unscaled delta increment predictions, and the bottom depicts true observations. Despite the very volatile chlorine concentrations (i.e.\ recurrent spiking), both models follow true values well.}
    \label{fig7.3}
    
\end{figure}

\autoref{fig7.3} depicts the teacher and student predictions for $\Delta_t$ and chlorine concentrations at a randomly sampled sensor over a randomly selected sequence. The sequence is from a randomly chosen validation scenario and ranges over 200 timesteps (\~ 16.67 hours). The model was trained given an HSR loss configuration.

\par
The random, varying magnitude of chlorine injections that were used to acquire these contexts are easily observable in \autoref{fig7.3}. Specifically, the spikes in concentration represent the worst-case sequences in which concentrations are rather noisy. Still, $T_{\phi}$ and $S_{\theta}$ follow the true observations and capture their variations well. The use of incremental predictions $\hat{\Delta}_t$ is similarly substantiated in \autoref{Delta_analysis}.

\subsection{Evolutionary Optimization}

Following the methods specified in Section~\ref{sc:method}, controllers were evolved to optimize the four WDS objectives. This section first presents quantitative results and compares to other methods, then evaluates different curriculum schemes, illustrates progress through Pareto fronts, and evaluates the effect of continual fine-tuning of the surrogate model.


\subsubsection{Quantitative Comparison} \label{qc}

\begin{table*}[t!]
    \centering
    \caption{Performance of baseline, evolved, and PPO controllers across the five evaluation objectives on the held-out scenario. Lower is better for all objectives. A knee-point solution in the Pareto front was used to represent the NSGA-II methods. The curricular NSGA-II agent captures the best tradeoffs between the possible ranges for all objectives.}
    \vspace*{-1.5ex}
    \begin{tabular}{lccccc}
    \toprule
    \textbf{Agent} & 
    \textbf{Bound} & 
    \textbf{Fairness} &
    \textbf{Smoothness} &
    \textbf{Cost} &
    \textbf{Infection Risk} \\
    \midrule
    
    \multicolumn{6}{l}{\textbf{Constant Injection}} \\
    \midrule
    0.3 mg/L     & 0.114 & \textbf{0.194} & --- & \textbf{1296.000} & 5.160 \\
    10 mg/L  & \textbf{0.105} & 0.196 & --- & 4.32 $\times$ 10$^4$ & 5.160 \\
    250 mg/L    & 0.840 & 8.915 & --- & 1.08 $\times$ 10$^6$ & 3.612 \\
    \midrule
    
    \multicolumn{6}{l}{\textbf{Random Injection}} \\
    \midrule
    $a_t \sim U(0, 250)$ & 0.408 & 4.289 & 87.024 & 5.334 $ \times$ 10$^5$ & 4.431 \\
    $a_t \sim U(0, 10000)$ & 16.673 & 185.722 & 3478.980 & 2.165 $ \times$ 10$^7$ & \textbf{0.106} \\
    \midrule
    
    \multicolumn{6}{l}{\textbf{Evolved Agents}} \\
    \midrule
    NEAT & 0.113 & 0.195 & \textbf{0.452} & 3456.000 & 5.160 \\
    NSGA-II & 0.401 & 24.014 & 17.693 & 7.500 $ \times$ 10$^5$ & 5.160  \\
    NSGA-II (\emph{Curriculum})  & 0.463 & 5.180 & 6.852 & 4.843 $ \times$ 10$^5$ & 3.627 \\
    \midrule
    
    \multicolumn{6}{l}{\textbf{PPO Agent}} \\
    \midrule
    PPO & 0.114 & 0.195 & 3.658 & 3636.568 & 5.160 \\
    \bottomrule
    \end{tabular}\\[2ex]
    
    \label{tab:agent_results}
    \vspace*{-1.75ex}
\end{table*}

\par
The evolved controllers were evaluated on a held-out three-day scenario, which was not utilized in training or validation of the surrogate or the prescriptors. Thus, the solutions were blind to any system-wide parameters governing fluid and reaction dynamics alongside the time/location of contamination events. 
\par
The results are summarized in \autoref{tab:agent_results}. The table includes five baseline policies: in the first three, chlorine was injected constantly at the rates of 0.3, 10, and 250 mg/L at each booster station. The lowest rate was chosen to fit within the boundary conditions for chlorine concentration, [0.2, 0.4] mg/L.  In the next two, chlorine was injected randomly as in initial surrogate training, sampling from $U(0, 250)$ and $U(0, 10000)$ mg/L. The latter is the maximum possible range for chlorine injections.
\par
NEAT controllers were evolved in three different ways (350 genomes for 100 generations; see \autoref{N-app} for details). The first was through NEAT's original single-objective evolutionary process. The four rewards were normalized and summed, acting as a single composite reward. Further, two NEAT controllers were evolved using NSGA-II. The first one did not utilize curriculum learning, and the second did. As a representative of these methods, an agent at the point of maximum curvature on the Pareto front (i.e.\ the knee point; ~\cite{Tang_Wang_Xiong_2023}) was identified and used for analysis. The details of these setups, including the resulting surrogate fine tuning, are in \autoref{S-app} and \autoref{N-app}.
\par
Finally, Proximal Policy Optimization (PPO) \cite{DBLP:journals/corr/SchulmanWDRK17} was used to train a controller with the same composite reward as the first NEAT controller. This approach took 2.71 M steps with separate LSTMs for the actor and critic networks. In addition, the model disabled orthogonal initialization, started with a log-standard-deviation of 1.0 for exploration, and trained with relatively short rollout steps (32) and a large batch size (256) to stabilize updates. This setup was found to be most effective in preliminary experiments.
\par
The first observation is that PPO, NEAT, and the small and medium constant injection policies (0.3, 10 mg/L) perform very similarly. It turned out that NEAT and PPO training collapsed, likely due to the noisy composite reward, resulting in very small injections ($< 10^{-2}$) at all booster stations. Thus, such off-the-shelf methods may not be sufficient in this task.

\par
The second observation is that the NSGA-II agents, and the NSGA-II curricular agent in particular, provided better tradeoffs for most objectives compared to random and constant injection policies. Compared to 250 mg/L, NSGA-II curricular had a similar infection risk despite a $\sim$ 40\% decrease in bound violations. Compared to $U(0, \ 250)$, it performed 12-13x better in injection smoothness with an 18\% decrease in infection risk despite similar bound violations. 

Third, the NSGA-II (\emph{Curricular}) agent dominated the non-curricular NSGA-II agent by a wide margin across all objectives except bound violations. Importantly, the curriculum improved infection risk by almost 30\%, alongside $\sim$ 5x and 4x decreases in fairness and smoothness.

\par
Thus, the evaluations in \autoref{tab:agent_results} demonstrate the advantage of NSGA-II over the baseline policies and PPO, and the curricular approach over the non-curricular approach.

\subsubsection{NSGA-II Objective Shaping}
To fully evaluate the efficacy of curriculum learning on NSGA-II, different ablations of the curriculum were evaluated. There are 12 different orderings: four maximally different ones were used for the evaluation, in addition to the default one used in Table~\ref{tab:agent_results}.  The same training and evolution framework was used as in Table~\ref{tab:agent_results}.

\begin{table*}[htbp]
    \centering
    \caption{Final objective values resulting from different curriculum ordering. Parentheses denote the initial two objectives, followed by the two remaining objectives in order.  Knee points in the Pareto front were used to represent each method; lower is better for all. The selection of the initial objective pair has a large effect, indicating synergistic or adversarial relationships between them.}
    \vspace*{-4ex}
    \begin{tabular}{lccccc}
        \hline
        \textbf{Ordering} & \textbf{Bound} & \textbf{Fairness} & \textbf{Smoothness} & \textbf{Cost} & \textbf{Inf. Risk} \\
        \hline
        (BC)SF & \textbf{0.154} & \textbf{1.195} & \textbf{5.277} & \textbf{1.224 $\times$ 10$^5$} & 5.14  \\
        (BC)FS & 0.246 & 3.267 & 6.598 & 2.783 $\times$ 10$^5$ & 5.079 \\
        (FC)BS & 0.615 & 3.864 & 11.045 & 7.215 $\times$ 10$^5$ & 3.794 \\
        (FC)SB & 0.422 & 5.432 & 7.978 & 3.792 $\times$ 10$^5$ & 3.855 \\
        (BF)SC & 0.463 & 5.180 & 6.852 & 4.843 $\times$ 10$^5$ & \textbf{3.627} \\
        \hline
    \end{tabular}
    \vspace{0.4em}\\
    {\footnotesize B=\textit{bounds}, C=\textit{cost}, F=\textit{fairness}, S=\textit{smoothness}}
    \\[1ex]
    \label{tab:nsga2_abl}
    \vspace*{-2.25ex}
\end{table*}

The results are summarized in \autoref{tab:nsga2_abl}. The first observation is that the intuition in Section~\ref{sc:curr_3} was indeed valid: the default (BF)SC order results in a good tradeoff with the best infection risk performance and midrange values in the other objectives. The second is that the order matters in a surprisingly systematic manner. In particular, there is quite a gap between agents that were initialized with \emph{bound violations} and \emph{cost control} (BC) as opposed to \emph{fairness} and \emph{cost control} (FC). The BC results look similar to the constant small injection policies (0.3 and 10 mg/L) and are likely due to the adversarial nature of bound violations and cost. Moreover, small injections mean that the total chlorine concentration largely depends on the reservoir's randomly initialized chlorine levels, which in turn means fewer deviations from fairness.

However, even though the (BC)SF approach minimizes four of the five objectives, it performs so poorly in infection risk that it would not be a good choice in practice. In contrast, the orderings that begin with FC, two synergistic objectives, result in more chlorine injection and are therefore able to reach lower infection risk.
\par
Together, these results demonstrate the nonlinear tradeoffs between objectives. Further work must be done to fully investigate the significance and variance of curricular ordering, and to devise methods to optimize it automatically for the domain. 

\subsubsection{Pareto Evolution}
Training dynamics for multi-objective problems is typically illustrated through Pareto fronts. 
NSGA-II specifically uses non-dominated sorting to build successive Pareto fronts that can be used to observe the evolution of the NEAT population in the objective landscape.

\begin{figure}
    \centering
    \begin{subfigure}{0.32\linewidth}
        \centering
        \hspace*{1ex}\includegraphics[width=\linewidth]{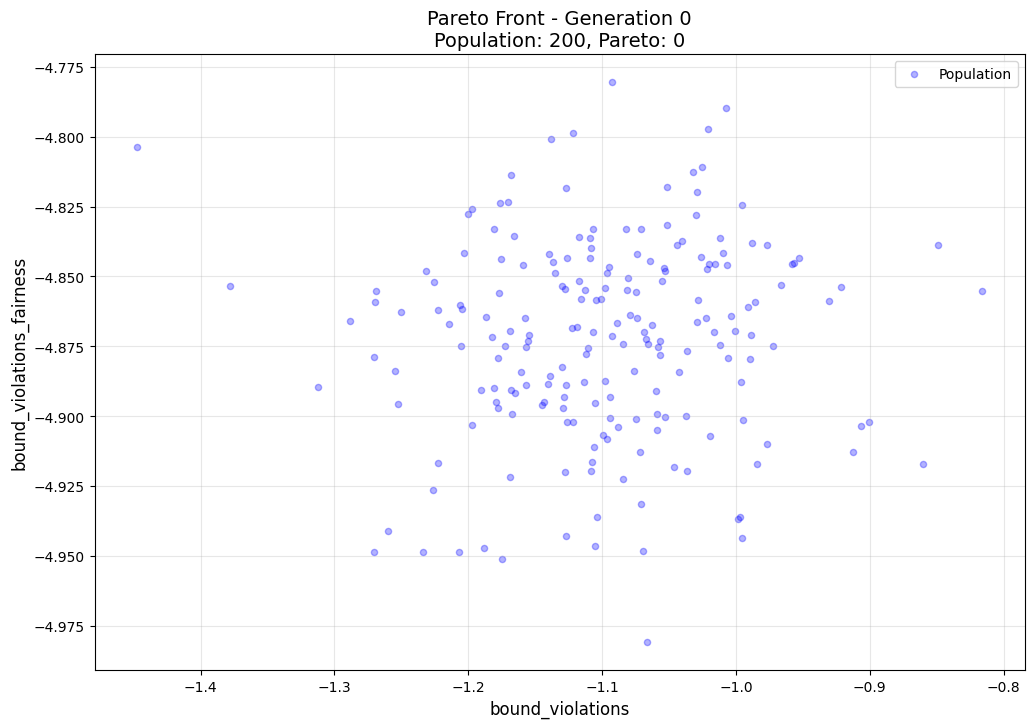}\\[-1ex]
        \caption{}
        \label{fig:g0}
    \end{subfigure}\\[1ex]
    \begin{subfigure}{\linewidth}
        \centering
        \includegraphics[width=\linewidth]{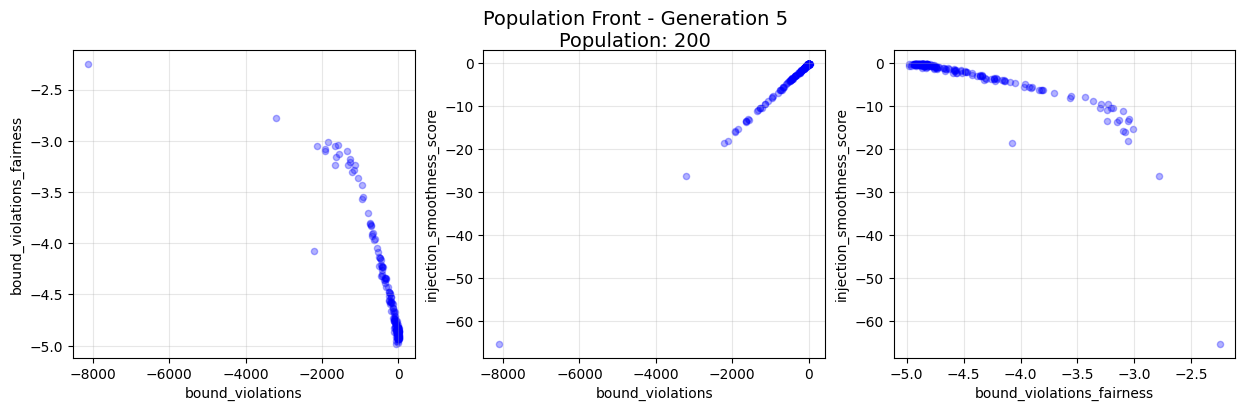}\\[-1ex]
        \caption{}
        \label{fig:g1}
    \end{subfigure}\\[1ex]
    \begin{subfigure}{\linewidth}
        \centering
        \includegraphics[width=\linewidth]{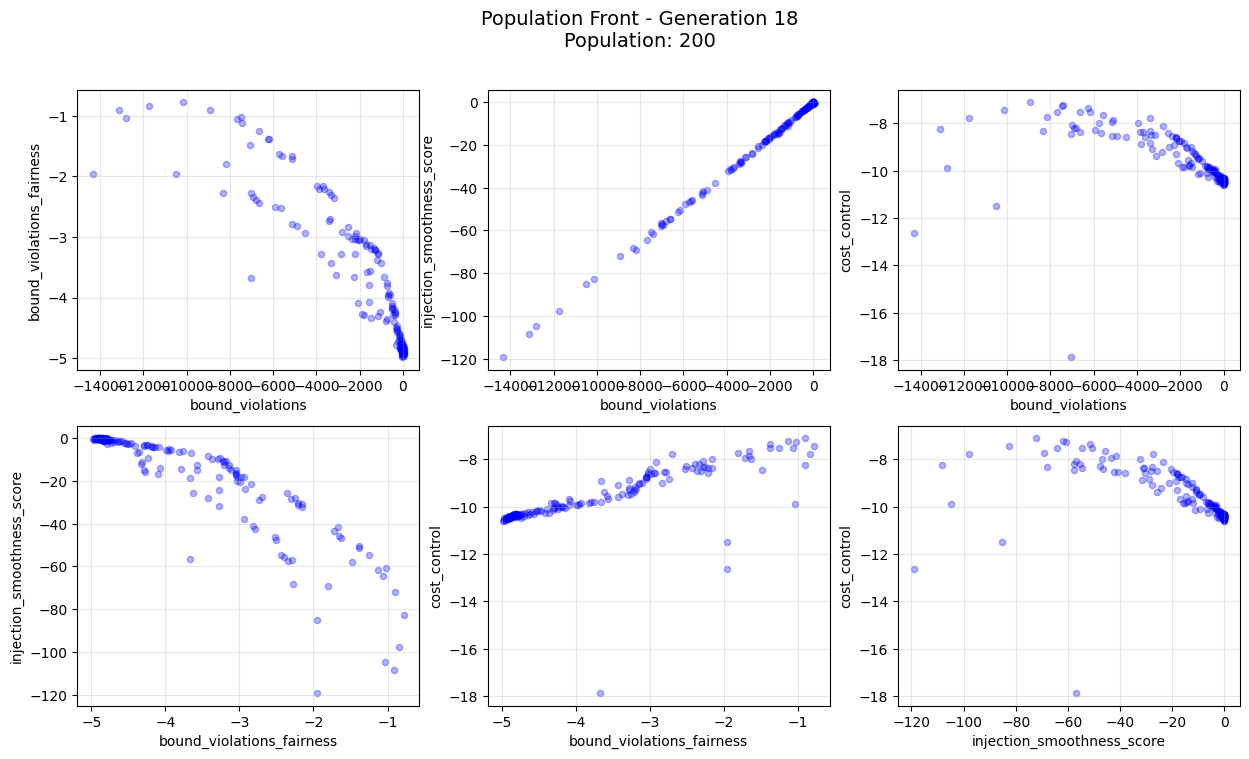}\\[-1ex]
        \caption{}
        \label{fig:g2}
    \end{subfigure}\\[-2.5ex]
    \caption{Pairwise population fronts with (a) two, (b) three, and (c) four objectives. The Pareto front evolves gradually as new objectives are added, and becomes fully formed when all objectives are included.}
    \label{fig:g3}
    \Description{Three figures depicting scatterplots of the population's pairwise evaluations against two, three, and four objectives. The first illustrates random evaluations (i.e.\ no front), the second shows a front but only covering half of the tradeoff space, and the last shows fully-formed Pareto fronts.}
\end{figure}

\par
At the start of evolution, there is no clear front due to random instantiation of network parameters (\autoref{fig:g0}). Optimization of the first two objectives (\emph{fairness} and \emph{bound violations}) followed by \emph{smoothness} converges the population into a Pareto front, shown in three pairwise graphs in \autoref{fig:g1}. The outliers in each graph illustrate the exploration done by NSGA-II and NEAT. 



\par
After optimizing the final objective, \emph{cost}, six pairwise fronts depicted in \autoref{fig:g2} result. The agents now cover the entire tradeoff region, offering a good selection of solutions.
Thus, the multi-objective nature of the algorithm is apparent, and the resulting Pareto front allows understanding of objective tradeoffs and makes it possible to select final agents for a variety of focuses.


\vspace*{-1.5ex}
\subsection{Effects of Surrogate Fine-tuning}

Given that the surrogate is initially trained using historical data that is independent of agent learning and curriculum, it was necessary to make it as general as possible. Otherwise, the evolution of agents may be muddled by artifacts of the surrogate, such as an inability to capture a certain range of the observation space. ESP mitigates this effect by continuously fine-tuning the surrogate on data generated by agents on real evaluations \cite{Francon_Gonzalez_Hodjat_Meyerson_Miikkulainen_Qiu_Shahrzad_2020}. 
\par
Continuing evolutionary search after surrogate fine-tuning causes NEAT to create many new species in the population. These species tend to occupy regions of the objective landscape that were previously unreachable under the prior surrogate, often crossing soft thresholds that had constrained earlier ESP iterations. 


\begin{figure}
    \centering
    \begin{subfigure}{0.48\linewidth}
        \centering
        \includegraphics[width=\linewidth]{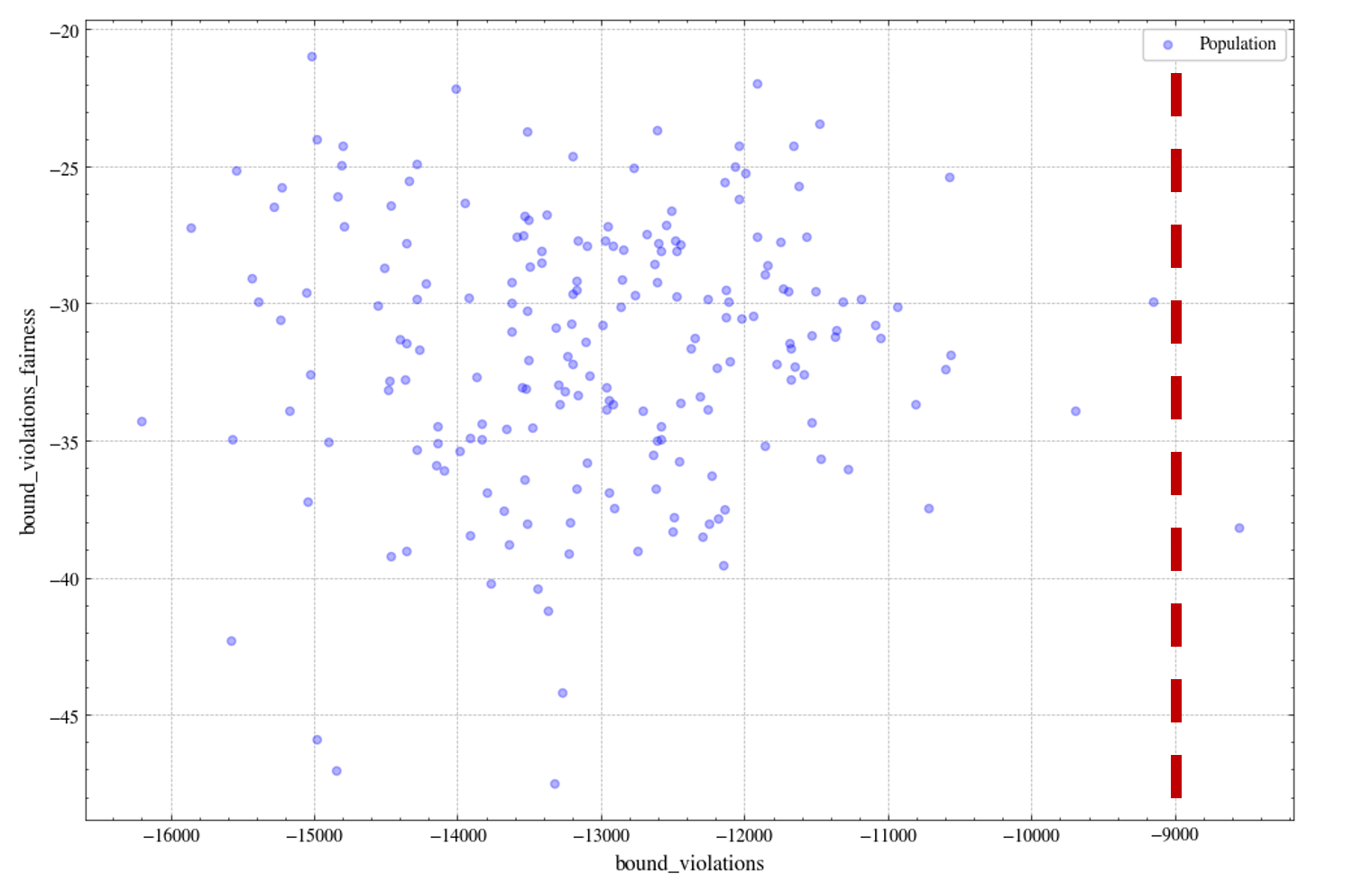}
        \caption{}
        \label{fig:side1}
    \end{subfigure}
    \hfill
    \begin{subfigure}{0.48\linewidth}
        \centering
        \includegraphics[width=\linewidth]{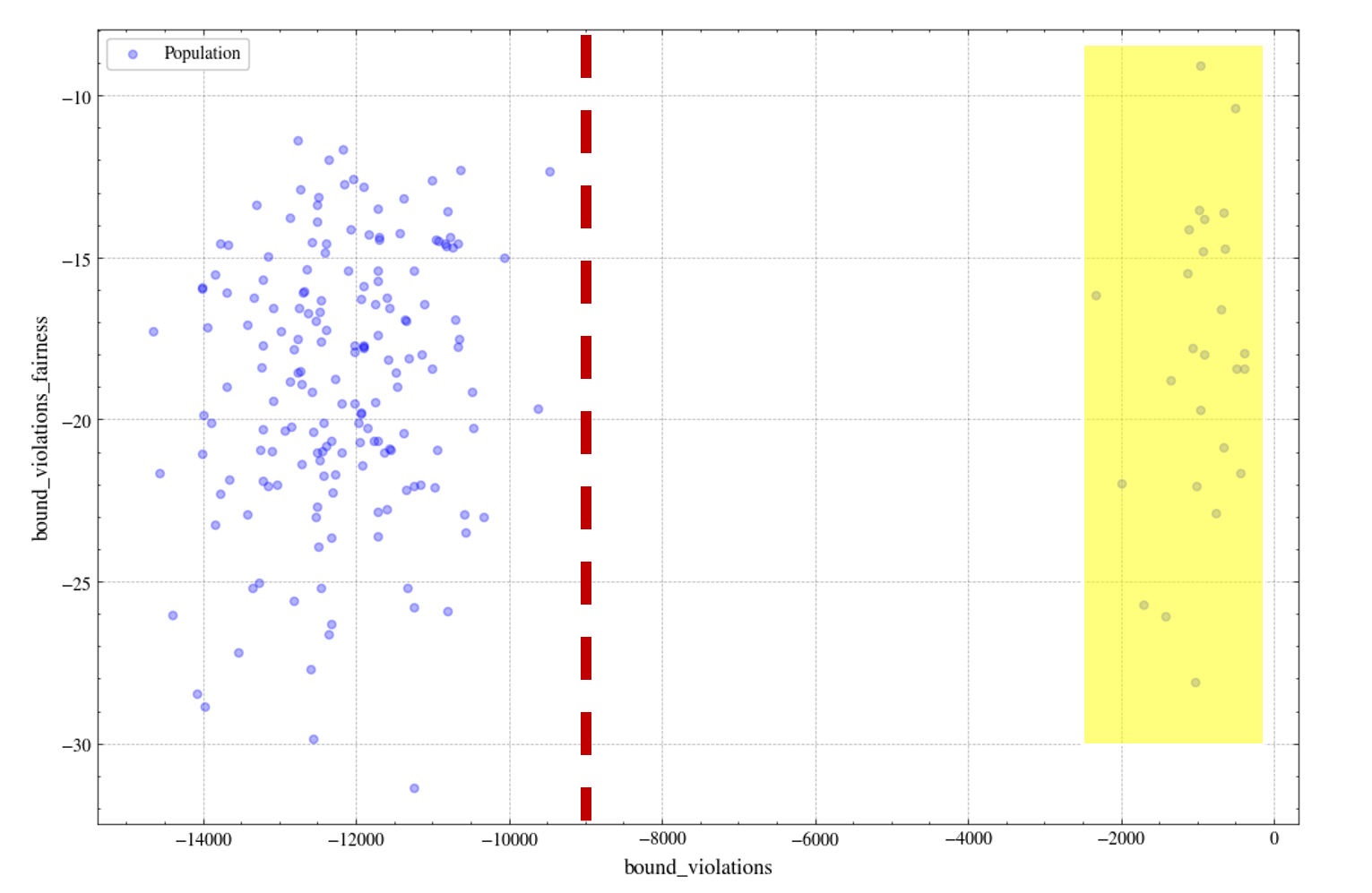}
        \caption{}
        \label{fig:side2}
    \end{subfigure}
    \vspace*{-2.5ex}
    \caption{NEAT innovation resulting from fine-tuning the surrogate model. The graphs illustrate the performance difference in bound violations between ESP iteration \textbf{(a)} zero to \textbf{(b)} one, with fine-tuning performed between them. Red lines indicate the level of the best solutions after iteration zero. After fine-tuning, a new species (highlighted in yellow) emerges above this level. In this manner, fine-tuning accelerates exploration of the objective landscape.}
    \label{fig:pareto_ft}
    \Description{Two Pareto fronts between cost and bound violations objectives at iterations 0 and 1. Almost all individuals (except one) at iteration 0 cannot surpass a reward of -10000 for bound violations, but after fine-tuning, a NEAT species occupies the reward landscape. That is, their evaluations show a better reward than -10000, and they can be distinguished in the Pareto front.}
\vspace*{-0.5ex}
\end{figure}

\par
This effect is seen in \autoref{fig:pareto_ft}. The species highlighted in yellow surpassed the previous performance in \emph{bound violations} ($y$) while simultaneously covering a breadth of the solutions in \emph{cost} ($x$).
\par
\begin{figure}[!tb]
    \centering
    \includegraphics[width=\linewidth]{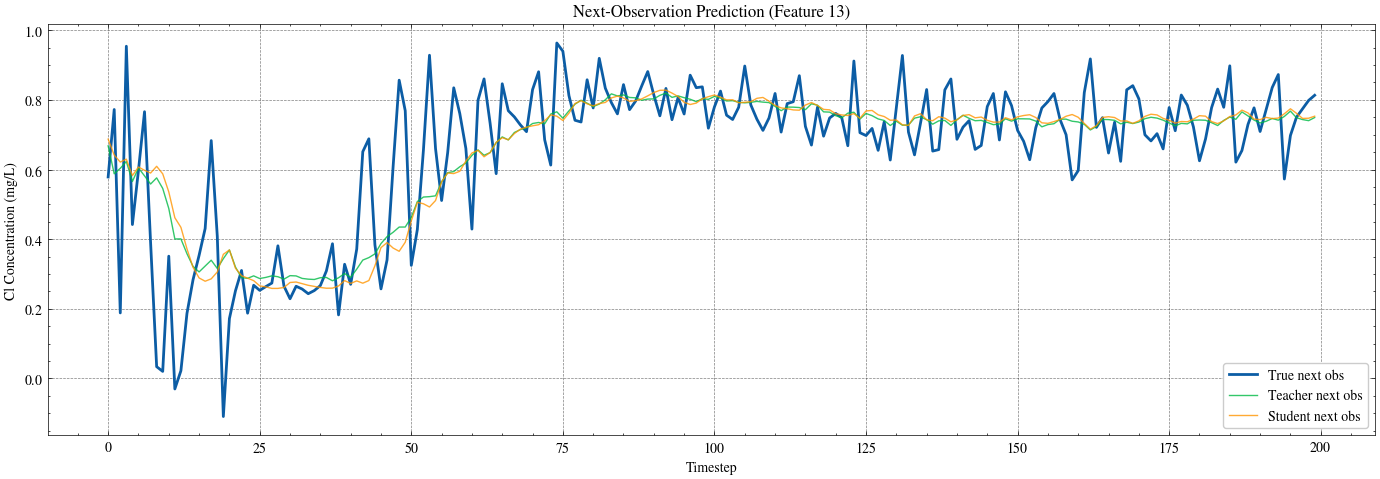}
    \caption{Predictions of $\hat{o}_{t+1}$ at ESP iteration four of the surrogate for chlorine sensor 13. Smooth student and teacher predictions illustrate the automatic regularization effect that co-optimization of the predictor and prescriptor has in ESP, resulting in faster progress and better solutions.}
    \label{fig12}
    \Description{A figure depicting teacher and student chlorine concentration predictions against the true value at ESP iteration four. While the true value is volatile, the teacher and student predictions are smooth and occur at the midpoints of the volatile spikes like a rolling-mean estimator.}
\vspace*{-2ex}
\end{figure}

In addition, after fine-tuning, the teacher and student surrogates produced single-step predictions that were substantially smoother than the ground truth. Compared to \autoref{fig7.3} in which $\hat{o}_{t+1}$ is jagged due to spikes in true chlorine concentrations, \autoref{fig12} shows smooth predictions. This smoothing arises from the synergistic co-optimization of the surrogate and prescriptor in which smoother state predictions allow the agent to inject chlorine more evenly across adjacent timesteps (thus improving the \emph{smoothness} objective). This observation is a new case of automatic regularization that emerges from the co-optimization of the predictor and prescriptor in ESP \cite{Francon_Gonzalez_Hodjat_Meyerson_Miikkulainen_Qiu_Shahrzad_2020}.

\vspace*{-2ex}
\section{Discussion and Future Work}

A new framework was proposed for optimizing WDS by leveraging surrogate modeling in ESP and curricular multi-objective optimization via NSGA-II. The resulting system controls the amount of chlorine present over time throughout the network given consumer demand, contamination events, reaction dynamics, and hydraulics.

The paper makes three main contributions: (1) Through numerical error metrics, the utility of knowledge distillation in producing causal, context-aware state-predictive models was corroborated. (2) The curricular NSGA-II approach was shown to provide better, centered tradeoffs for most objectives compared to random and constant injection policies and the PPO policy-gradient method. (3) Surrogate fine-tuning in ESP was demonstrated to result in enhanced innovation through speciation pressure, and automatic regularization through co-optimization of predictors and prescriptors. These results suggest that evolutionary methods can be used to optimize WDS both in digital twins and the real-world.
\par
There are several avenues of future work. The lack of longer EPANET simulations in the training scheme, such as on the order of months or years, makes it difficult for the surrogate to generalize over a range of real-world contamination events, such as the introduction of an over-concentrated pathogen. Running larger experiments, i.e.\ with a larger population and more ESP generations, may unveil new phenomena inherent to the framework and provide better objective evaluations. Third, the approach needs to be evaluated in a wider variety of network structures, potentially identifying aspects of the framework that are already general and aspects that need to be customized. Fourth, the framework can be applied to other spatiotemporal optimization problems, such as those in transportation, communication, and construction.

\vspace*{-1ex}
\section{Conclusion}
This paper describes a new framework in optimizing chlorination control in water distribution systems: curricular multiobjective neuroevolution with a fine-tuned surrogate model. Starting with the simplest objectives, more challenging objectives are included step by step, resulting in a comprehensive Pareto front. The surrogate model is fine-tuned after each evolutionary iteration, promoting innovations and regularization. A transparent set of trade-offs is created, enabling decision makers to select policies based on local resources, constraints, and priorities. The framework can be applied to other spatiotemporal optimization problems as well, thus potentially contributing to sustainable growth and resilience of critical infrastructure around the world.


\newpage
\bibliographystyle{ACM-Reference-Format}
\bibliography{main}

\appendix
\clearpage\clearpage
\section{Reward Engineering} \label{R-app}
To evolve agents with NSGA-II, it is necessary to quantify the performance of each agent according to each objective. In turn, these objectives must be measured by a numerical value $\omega_i \in R$ where larger (or smaller) values indicate worse performance, or violations of the objective. Scale differences in the rewards typically do not matter in NSGA-II due to its focus on non-dominated sorting and ranking.
\par
Typically, the quantification of these rewards falls into the broad field of reward engineering in RL, which is crucial to guiding agent learning and/or evolution \cite{ibrahim2024comprehensiveoverviewrewardengineering}. The reward functions in this paper are a practical initial formulation; they may be further refined in future work.
\par
The objectives/rewards at each timestep $t$ were calculated as follows. Let the predicted chlorine concentrations at $|\mathbf{o}_t| = n$ monitored nodes be $\mathbf{o} = (o_1, \dots, o_n) \in \mathbb{R}_+^{17}$.
Then:

\paragraph{Bound Violations.}
The single-most important objective is ensuring chlorine concentrations at all sensor nodes are between [0.2, 0.4] mg/L. To do so, define violations above and below the allowable limits as
\[
\mathrm{over}_i = \max(o_i - o_{\max},\, 0), \qquad
\mathrm{under}_i = \max(o_{\min} - o_i,\, 0),
\]
where $o_{\mathrm{min}} = 0.2$ and $o_{\mathrm{max}} = 0.4$. The overall objective includes both penalties based on violations of these bounds ($R_{\mathrm{raw}}$) and a compliance bonus to shape rewards and to make them less sparse ($B$). Importantly, the $under$ bound violations are weighted five times as heavily as the $over$ bound violations due to observed convergence of agents at local optima in which almost no chlorination was injected into the WDS. Thus,
\begin{equation*}
    \begin{gathered}
            R_{\mathrm{raw}}
        = \frac{10}{n}\sum_{i=1}^{n}\!\left(\mathrm{over}_i + 5\,\mathrm{under}_i\right),
        \qquad
        B = \frac{-5}{n}\sum_{i=1}^{n}\mathbf{1}_{\{x_{\min}\le x_i\le x_{\max}\}},\\[2mm]
        R_{\mathrm{bounds}} = 50\,\tanh\!\left(\dfrac{R_{\mathrm{raw}}+B}{50}\right).
    \end{gathered}
\end{equation*}
The overall reward, $R_{\mathrm{bounds}}$, normalizes these two terms and trivially scales the reward by 50 solely for readability.

\paragraph{Spatial Fairness.} The agent must also ensure that chlorine concentrations over $\mathbf{o}$ are even over all consumer nodes. Define the node-wise deviation score as
\[
s_i =
\begin{cases}
o_i - o_{\max}, & o_i > o_{\max},\\[1mm]
o_{\min} - o_i, & o_i < o_{\min},\\[1mm]
0, & \text{otherwise}.
\end{cases}
\]
To ascertain deviation of each node over the mean, define $\bar{s} = \frac{1}{n}\sum_{i=1}^n s_i$. Thus, the fairness penalty is
\[
R_{\mathrm{fair}}
= \frac{\max_i s_i - \min_i s_i}{\bar{s} + 10^{-6}},
\]
where $10^{-6}$ ensures a valid value.

\paragraph{Smoothness.}
Per-node chlorine injections across adjacent timesteps must also be similar to ensure proper functioning of booster stations and, implicitly, spatially even chlorine concentrations. Let the current injection action vector be $\mathbf{a}_t = (a_{t,1},\dots,a_{t,m})$
and the previous actions $\mathbf{a}_{t-1}$. Then, the reward is

\[
R_{\mathrm{smooth}} =
\begin{cases}
\dfrac{1}{m}\displaystyle\sum_{j=1}^{m} \left| a_{t,j} - a_{t-1,j} \right|, & t > 0, \\[4mm]
0.1, & t = 0~,
\end{cases}
\]
where a small constant penalty is applied to the first timestep $t = 0$.

\paragraph{Cost Control.}
A ubiquitous objective is the minimization of resource expenditure. In this case, the bulk amount of chlorine that is injected by the policy must be minimized. Chemical usage is simply penalized via the sum of injections:
\[
R_{\mathrm{cost}}
= \sum_{j=1}^m a_{t,j}.
\]

\paragraph{Infection Risk.}
Lastly, infection risk estimates the probability of a simulated individual falling sick with a water-borne disease. This estimate is typically calculated via the Quantitative Microbial Risk Assessment (QMRA) framework and dose-response model \cite{aichall, Clements_Crank_Nerenberg_Atkinson_Gerrity_Hannoun_2024}. 
This calculation is only possible outside of the simulation due to the randomness of contamination events occurring. As such, there may or may not be contamination events in some training contexts.

\paragraph{Reward Vector.}
The vector
$\mathbf{R} = \big(
R_{\mathrm{bounds}},\;
R_{\mathrm{fair}},\;
R_{\mathrm{smooth}},\;
R_{\mathrm{cost}}
\big)$ was passed to the NSGA-II state as $-\mathbf{R}$ to fit a maximization scheme in which large, negative rewards indicate increasing non-compliance. Note that infection risk was not included in this vector (for the above reasons). However, it is implicitly optimized as a side effect of optimizing the other objectives. As infection risk is inherent to contamination events, agents with efficient chlorination adhering to the fairness, smoothness, and bound violations objectives are necessary to mitigate possible infection.

\section{Surrogate Training} \label{S-app}
The surrogate modeling and the teacher-student architecture are described in detail in this section. Moreover, specifics about surrogate training and data acquisition are expounded upon.

\subsection{Problem Definition}
The predictor is a surrogate that approximates the
one-step hydraulic and water quality state in the EPANET-based simulator. Let $t \in \{1,\dots,T\}$ denote discrete simulation time
(5-minute intervals).  
At each time step, the system state is represented as
\begin{equation}
x_t = \big[\, f_t,\; o_t,\; a_t \,\big],
\end{equation}
where  
$f_t \in \mathbb{R}^{2}$ are hydraulic features (flows),  
$o_t \in \mathbb{R}^{17}$ are water quality observations (chlorine concentrations),  
and $a_t \in \mathbb{R}^{5}$ is the injected chlorine action vector. In the EPANET simulation, the former two are calculated by the hydraulic emulator, and the latter is predicted by the control system. In this case, while the agent still predicts $a_t$, $f_t$ and $o_t$ are modeled by the surrogate, say $g_{\theta}(\cdot)$.
\par
To enable such state prediction, a uni-directional LSTM architecture  \cite{lstm} is used for the surrogate. The dynamics of a WDS are inherently non-linear and exhibit strong, delayed temporal dependencies. The LSTM input consists of context vectors that capture the time-based dynamics of both hydraulics and water quality.
\par
Given a context window of length/horizon $H$
\begin{equation}
C_t = \{ x_{t-H+1},\, x_{t-H+2},\, \dots,\, x_t \},
\end{equation}
the surrogate model predicts the next-step hydraulic and water-quality
variables,
\begin{equation}
\hat{y}_{t+1} = g_\theta(C_t)
\qquad\text{where}\qquad
y_{t+1} = [\, f_{t+1},\, o_{t+1} \,].
\end{equation}
\par
To generate predictions over an arbitrary number of timesteps, the surrogate is unrolled autoregressively to produce multi-step predictions:
\begin{align}
    \hat{x}_{t+1} &= g_\theta(C_t), \\
    \hat{x}_{t+2} &= g_\theta\!\left(\{x_{t-H+2}, \ldots, x_t,\, \hat{x}_{t+1}\}\right), \\
    &\;\;\vdots \nonumber
\end{align}
where at each step the newest predicted state replaces the oldest element in the context window. 
\par
Furthermore, instead of predicting the true values of $f_t$ or $o_t$, the surrogate is trained to predict $\Delta_t = [\Delta f_t, \ \Delta o_t]$, or the change in the flows and concentrations. Typically, the magnitude of these measurements varies widely based on parameters governing the hydraulic and water quality transformations of a scenario. Still, demands remain similar, leading to similar usage patterns. As such, modeling the changes tends to be smoother and accumulate less error over autoregressively-unrolled predictions of the WDS state. The model predicts this change as
\begin{equation}
\Delta_t = y_{t+1} - y_t,
\qquad
\hat{\Delta}_t = g_\theta(C_t), 
\label{delta:eq}
\end{equation}
and reconstructs the next observation as $\hat{y}_{t+1} = y_t +
\hat{\Delta}_t$. This method is used to train the LSTM to predict single steps auto-regressively (with $\hat{y}_t$ instead of $y_t$). The predictions focus on chlorination readings $o_t$ rather than flows $f_t$ in order to take advantage of more observations (17 vs. two).


\subsection{Teacher--Student Architecture}
A knowledge-distillation framework is adopted to train the surrogate.  
The \emph{teacher} is a bidirectional LSTM (BiLSTM) that has access to
both past and limited future context \cite{huang2015bidirectionallstmcrfmodelssequence}.  
For each time index $t$, the teacher receives a window
\begin{equation}
\widetilde{C}_t
= \{ x_{t-H+1},\dots,x_t,\dots,x_{t+K} \},
\end{equation}
where $K$ is a short lookahead (in this case, $H = K$).  
In turn, the teacher produces highly accurate but non-causal predictions:
\begin{equation}
T_\phi(\widetilde{C}_t) = \hat{\Delta}^{(T)}_t.
\end{equation}
Because the bidirectional model requires a look-ahead in time, it is not realistic and cannot be used online in either the evolution of prescriptors or the modeling of WDS states. However, it can be used to generate training targets for another model that can fill that role, i.e.\ it can serve as a \emph{teacher}. This setup is known as knowledge distillation, in which the outputs of a larger, pre-trained model are used to guide training of a smaller model \cite{hinton2015distillingknowledgeneuralnetwork}. 
Therefore, the \emph{student} is a causal LSTM that only receives past context,
\begin{equation}
S_\theta(C_t) = \hat{\Delta}^{(S)}_t,
\end{equation}
and therefore can be used during real-time control and evolution of agents.  
The goal is to train the student to approximate the teacher while still fitting the true EPANET dynamics. These dual training objectives are achieved through different loss terms during distillation.

\subsubsection{Training Objective}
The teacher model is trained via MSE loss. On the other hand, 
the student model is trained with a hybrid distillation loss consisting of several terms \cite{Kim_Oh_Kim_Cho_Yun_2021}:
\begin{multline}
    \mathcal{L}(\theta)
=
\lambda_{\mathrm{hard}}
\underbrace{\|\hat{\Delta}^{(S)}_t - \Delta_t\|_2^2}_{\text{ground-truth loss}} \
+ \
\lambda_{\mathrm{soft}}
\underbrace{\|\hat{\Delta}^{(S)}_t - \hat{\Delta}^{(T)}_t\|_2^2}_{\text{teacher-loss}} \\
+ \lambda_{\mathrm{feat}}\mathcal{L}_{feat} + \lambda_{\mathrm{rollout}}\mathcal{L}_{\mathrm{rollout}},
\end{multline}

\noindent where the $\lambda$ values are coefficients to weigh each term. The first term (\emph{hard loss}) enforces fidelity to the EPANET
simulation. The second term (\emph{soft loss}) distills knowledge from the high-capacity, non-causal teacher. These two are the more consequential terms to enable knowledge-distillation. In the experiments, two variants were also evaluated:
\begin{align}
\mathcal{L}_{\text{feat}}
&=
\left\|
h^{(S)}_t - h^{(T)}_t
\right\|_2^2 \label{feat} \\
\mathcal{L}_{\text{rollout}}
&=
\sum_{k=1}^{K}\gamma^k
\left\|
\hat{x}_{t+k} - x_{t+k}
\right\|_2^2 .\label{rollout}
\end{align}

\autoref{feat} (\emph{feature loss}) tries to align the feature-level representation of the teacher and student using their final hidden-state activations, $h^{(S)}_t$ and $h^{(T)}_t$. \autoref{rollout} (\emph {rollout loss}) penalizes accumulation of error over unrolled horizons $K$. Just like in typical RL, a discounting factor $\gamma \in [0,1]$ is applied to more heavily weigh timesteps closer to $t$ rather than $t + K$. Different coefficients were evaluated for these loss terms in \autoref{lfabl} in order to utilize the best combination in distilling the teacher to the student model.

To ensure numerical stability of the observations and actions over all generalized scenarios, $o_t$, $f_t$, and $a_t$ are independently normalized to induce a normal distribution centered around $\mu = 0$ with $\sigma = 1$. Moreover, $\Delta_t$ is normalized similarly. Predictions by the surrogate were transformed into $\hat{y}_{t+1}$ by first unscaling and then adding $\hat{\Delta}_t$ to $y_t$. 

\subsubsection{Data Processing}
The EPANET simulator is run under a variety of hydraulic and demand
scenarios, producing trajectories, $\{ (f_t, o_t, a_t, o_{t+1}) \}_{t=1}^T.$ Nine three-day scenarios are utilized for the proposed scheme. To collect initial trajectories to train the teacher and student models, each scenario is simulated with uniform sampling of $[0, 1000]$ mg/L of chlorine as actions. Data collection is repeated 10 times per scenario, resulting in 90 distinct scenarios used for training. Context samples are constructed from this data. Teacher inputs are of shape $(H+K,\,F+O+A)$ while the uni-directional student only takes contexts that enforce causality, $(H,\;F+O+A)$ (i.e.\ no lookahead) where $F+O+A = 24$ and $H = K = 12$. Therefore, the teacher utilizes information from the past and future one hour, whereas the student utilizes the state from only the past hour.
\par
To ensure generalization across operating conditions, the dataset is
split by scenario ID: 10\% of scenario contexts are held out entirely for
validation. All results are calculated on these validation scenarios.

\subsubsection{Training Parameters.}
Both teacher and student models were trained for 10 ESP iterations. Per-iteration evolution of NEAT agents was run for 50 generations with 350 individuals. The non-curricular agent was optimized on the four objectives for all 10 iterations. On the other hand, the curricular agents were trained on \emph{bound violations} and \emph{fairness} for two ESP iterations, then \emph{smoothness} and \emph{cost control} were included in the scheme every two ESP iterations, in that order. Therefore, the \emph{smoothness} objective was introduced in iteration three while the \emph{cost control} objective was integrated in iteration five. At the end of evolution, the five best genomes were selected according to Pareto-optimality and evaluated on five randomly sampled scenarios. The trajectories collected from each evaluation were used to fine-tune $T_{\phi}$ for 20 epochs. Then, $S_{\theta}$ was fine-tuned via distillation using the HSR configuration (see \autoref{lfabl}) for 10 epochs. 

\par
Initial surrogate training was done over 50 epochs for both student and teacher models. Early stopping was implemented on validation loss for the teacher, and the best models were returned with respect to validation loss for both models. 

\begin{table}
    \centering
    \caption{Surrogate model architectures and hyperparameters.}
    \begin{tabular*}{\linewidth}{lcc}
        \hline
        \textbf{Hyperparameter} & \textbf{Teacher} ($T_{\phi}$) & \textbf{Student} ($S_{\theta}$) \\
        \hline
        Architecture & 2-layer BiLSTM & 1-layer LSTM \\
        Hidden size & 128 per direction & 64 \\
        Dropout & 0.1 & 0 \\
        Context length & 24 steps & 12 steps \\
        Output target & $[\Delta f_t, \ \Delta o_t]$ & $[\Delta f_t, \ \Delta o_t]$ \\
        Optimizer & Adam & Adam \\
        Learning rate & $1\times 10^{-3}$ (cosine decay) & $5\times 10^{-4}$ \\
        Batch size & 256 & 256 \\
        L2 Penalty & $1\times 10^{-5}$ & $1\times 10^{-6}$ \\
        Training epochs & 50 (early stopping) & 50 \\
        Fine-tuning epochs & 20 (early stopping) & 10 \\
        \hline
    \end{tabular*}\\[2ex]
    \label{tab:surrogate_hparams}
\end{table}

\subsubsection{HSR configuration analysis.}

\begin{figure}
    \centering
    \begin{subfigure}{0.48\linewidth}
        \centering
        \includegraphics[width=\linewidth]{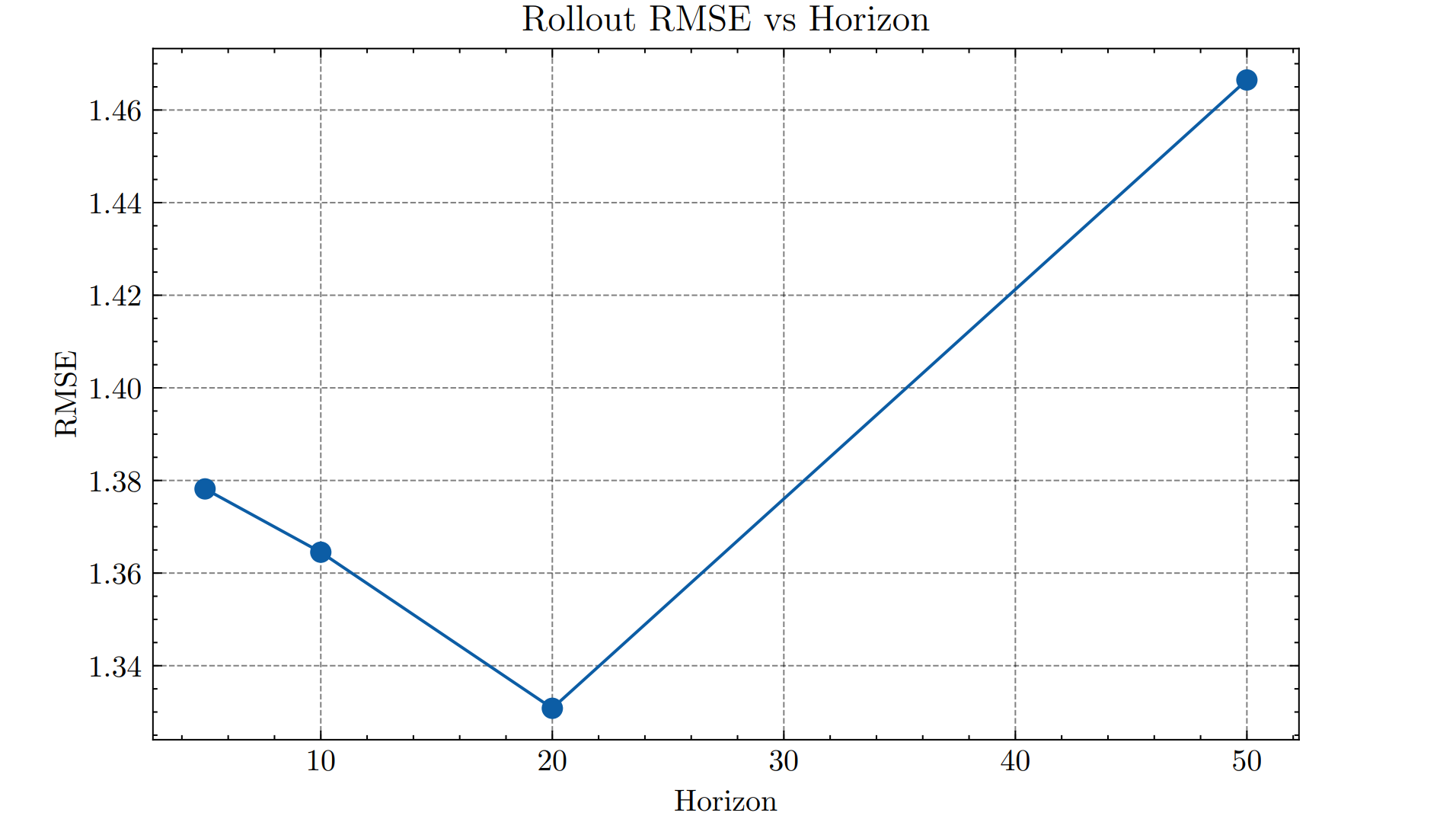}
        \caption{}
        \label{fig:fig_rmse}
    \end{subfigure}
    \hfill
    \begin{subfigure}{0.5\linewidth}
        \centering
        \hspace*{-5ex}\includegraphics[width=1.2\linewidth]{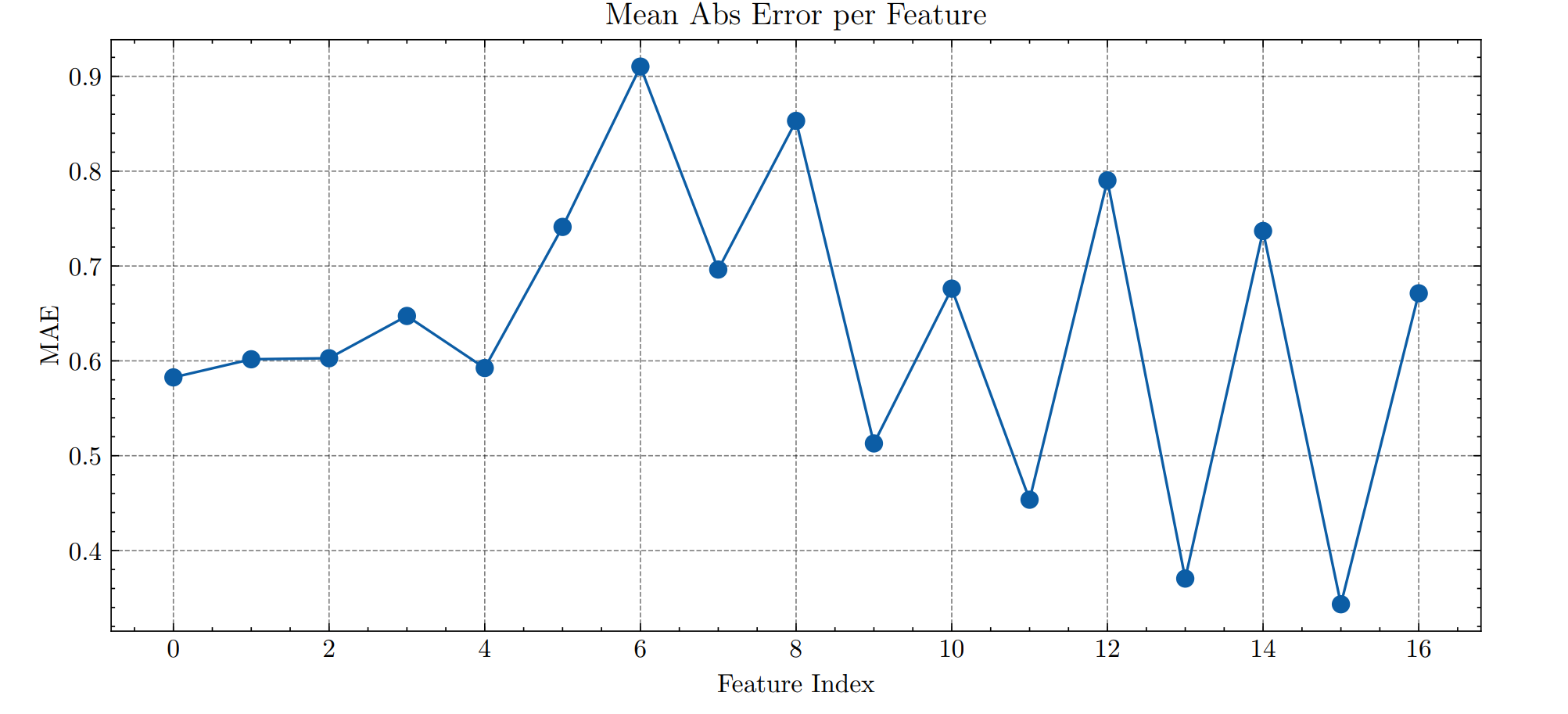}
        \caption{}
        \label{fig:fig_mae}
    \end{subfigure}

    \caption{Accuracy of HSR across time and space in single-step prediction. (a) RMSE across four different prediction horizons. (b) MAE across the different sensors. Rollout optimization benefits lookahead within the context window size, while some sensors are inherently harder to model than others.}
    \label{fig:fig_rmse+mae}
    \Description{Two figures showing rollout RMSE and MAE per feature. The rollout RMSE figure shows a decrease in RMSE from 2 to 10 and 20 timesteps, then an increase from 20 to 50 timesteps. The MAE figure shows relatively constant MAE for features 0 to 4 and volatile MAE for features 5 to 16.}
\end{figure}

As the prediction horizon increases, the validation loss first decreases from five to 20 timesteps, and then increases at 50 timesteps (\autoref{fig:fig_rmse}). This increase likely originates from the complexity of chlorine's temporal degradation, which occurs over the order of minutes and hours. 

\par
In \autoref{fig:fig_mae}, the spatial variation in chlorine concentration is apparent in the significant differences between MAE for different sensing nodes. 
Sensors nearby (e.g.\ 0-4) have similar MAE, whereas those further apart (6-16) vary significantly. 
In principle, it may be possible to normalize the outputs of each sensor, but this variation across sensors would still remain: there are always a few sensors whose values are hard to predict.

\subsubsection{Efficacy of $\Delta_t$ for surrogate training.}
\label{Delta_analysis}
\begin{figure}
    \centering

    \begin{subfigure}{\linewidth}
        \centering
        \includegraphics[width=\linewidth]{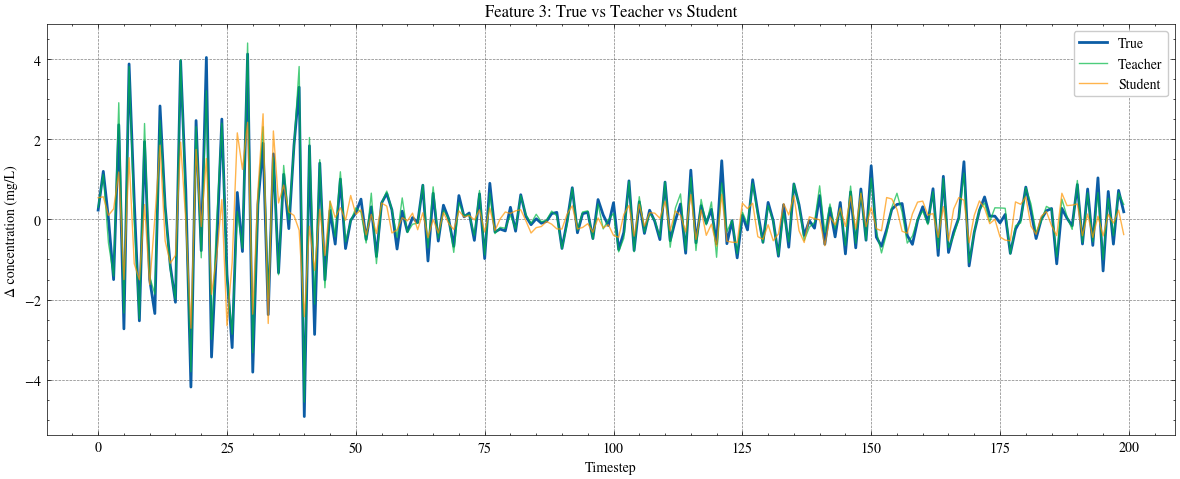}
        \caption{}
        \label{fig7.1}
    \end{subfigure}
    \hfill
    \begin{subfigure}{\linewidth}
        \centering
        \includegraphics[width=\linewidth]{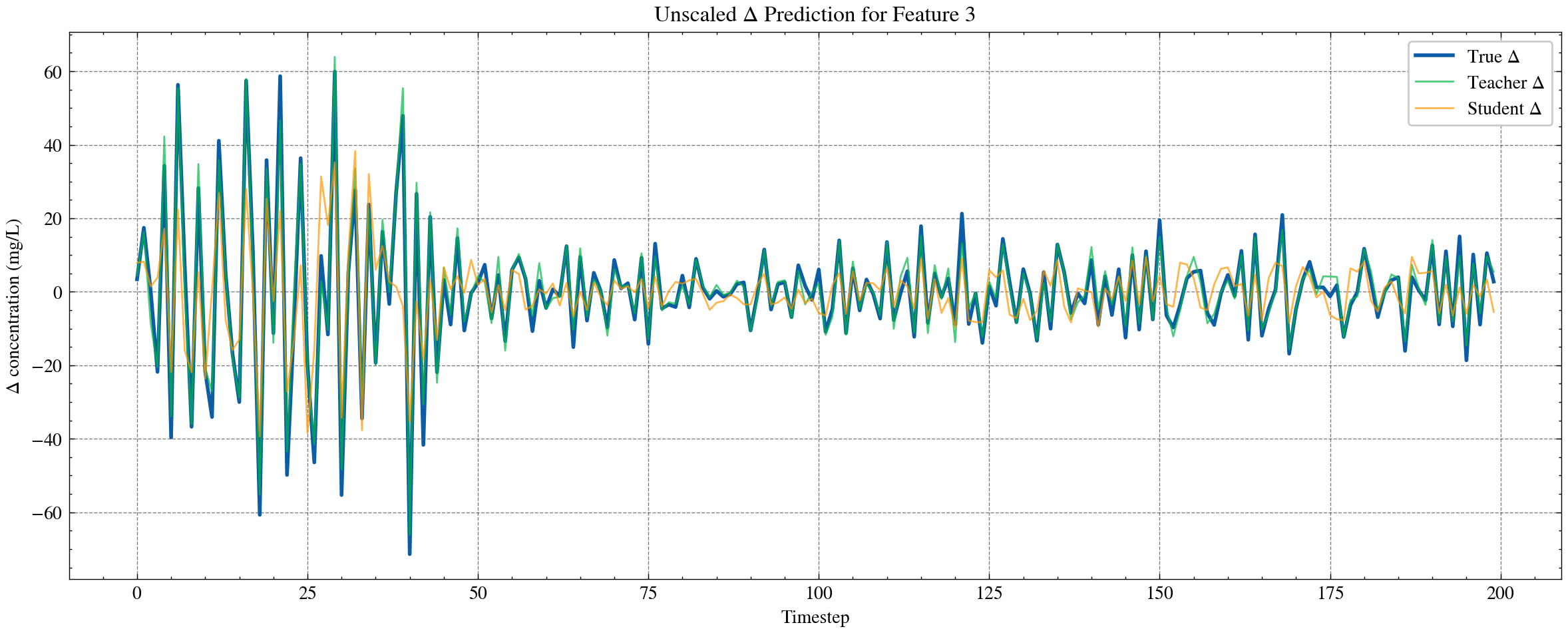}
        \caption{}
        \label{fig7.2}
    \end{subfigure}
    \caption{Comparison of $T_{\phi}$ vs. $S_{\theta}$ vs. True values for \textbf{a)} scaled $\Delta_t$ and \textbf{b)} unscaled $\Delta_t$. Low error in $S_{\theta}$ predictions indicates that the composition of scaling with incremental predictions enables accurate modeling of the hydraulic state.}
    \label{fig7}
    \Description{2 figures illustrating the predictions of the teacher and student model against the true chlorine concentration. The top figure follows scaled delta increment predictions, and the bottom shows unscaled delta increment predictions, and the bottom depicts true observations. In both instances, the models follow the volatile chlorine concentrations well, indicating that scaling and the use of delta increments are beneficial to capturing the volatile state.}
\end{figure}

The $\hat{\Delta}_t$ follows $\Delta_t$ particularly closely in \autoref{fig7.1} and \autoref{fig7.2}. More errors can be observed, especially for the student model in \autoref{fig7.3}, but its predictions follow $y_t$ close enough to conclude that it has learned the fundamentals of the physical constraints dictating the system.

\section{NEAT Configuration} \label{N-app}

Initial NEAT configuration files for experiments are self-contained in \textit{.config} files in the \texttt{neat-python} library \cite{neat-python}. NEAT-only training used the same configuration except for NSGA-II-specific settings at the bottom of the file. The configuration file with all settings used is listed in \autoref{fig:config_side_by_side}.

\begin{figure*}[htbp]
\centering

\begin{minipage}[t]{0.48\textwidth}
\begin{mdframed}[
    linewidth=0.5pt,
    roundcorner=2pt,
    innerleftmargin=10pt,
    innerrightmargin=10pt,
    innertopmargin=10pt,
    innerbottommargin=10pt
]
\scriptsize
\ttfamily
\begin{verbatim}
[NEAT]
fitness_criterion     = max
fitness_threshold     = 10000 # arbitrary
pop_size              = 350
reset_on_extinction   = False

[DefaultGenome]
# node activation options
activation_default      = relu
activation_mutate_rate  = 0.1
activation_options      = relu softplus

# node aggregation options
aggregation_default     = sum
aggregation_mutate_rate = 0.1
aggregation_options     = sum

# node bias options
bias_init_mean          = 0.0
bias_init_stdev         = 2.0
bias_max_value          = 1000
bias_min_value          = 0
bias_mutate_power       = 0.5
bias_mutate_rate        = 0.7
bias_replace_rate       = 0.1

# genome compatibility options
compatibility_disjoint_coefficient = 1.0
compatibility_weight_coefficient   = 0.5

# connection add/remove rates
conn_add_prob           = 0.5
conn_delete_prob        = 0.5

# connection enable options
enabled_default         = True
enabled_mutate_rate     = 0.1

feed_forward            = False
initial_connection      = partial_direct 0.3
\end{verbatim}
\end{mdframed}
\end{minipage}
\hfill
\begin{minipage}[t]{0.48\textwidth}
\begin{mdframed}[
    linewidth=0.5pt,
    roundcorner=2pt,
    innerleftmargin=10pt,
    innerrightmargin=10pt,
    innertopmargin=10pt,
    innerbottommargin=10pt
]
\scriptsize 
\ttfamily
\begin{verbatim}
# node add/remove rates
node_add_prob           = 0.5
node_delete_prob        = 0.2

# network parameters
num_hidden              = 0
num_inputs              = 17
num_outputs             = 5

# node response options
response_init_mean      = 1.0
response_init_stdev     = 0.0
response_max_value      = 30.0
response_min_value      = 1.0
response_mutate_power   = 0.0
response_mutate_rate    = 0.0
response_replace_rate   = 0.0

# connection weight options
weight_init_mean        = 1.0
weight_init_stdev       = 2.0
weight_max_value        = 1000
weight_min_value        = 0
weight_mutate_power     = 0.5
weight_mutate_rate      = 0.8
weight_replace_rate     = 0.1

[DefaultSpeciesSet]
compatibility_threshold = 3.0

[DefaultStagnation]
species_fitness_func = max
max_stagnation       = 20
species_elitism      = 10

[NSGA2Reproduction]
survival_threshold = 0.2

\end{verbatim}
\end{mdframed}
\end{minipage}

\caption{Configuration file for NEAT experiments via NSGA-II using \texttt{neat-python} library.}
\label{fig:config_side_by_side}
\Description{Configuration file for NEAT algorithm, which can be directly plugged in as a configuration file for the neat-python library.}
\end{figure*}

\end{document}